\documentclass[journal]{IEEEtran}

\usepackage{cite}

\usepackage{amsmath}

\usepackage{amssymb}  

 \usepackage{array}

\usepackage{fixltx2e}

\usepackage{url}

\newcommand{\Rmnum}[1]{\uppercase\expandafter{\romannumeral #1}}
\usepackage{color}

\usepackage{multirow}

\usepackage{threeparttable}

\newcommand{\tabincell}[2]{\begin{tabular}{@{}#1@{}}#2\end{tabular}}

\usepackage{graphics} 
\usepackage{epstopdf}
 \usepackage{epsfig} 

\begin{document}

\title{Statistical Pattern Recognition for Driving Styles Based on Bayesian Probability and \\ Kernel Density Estimation }

\author{{Wenshuo Wang, Junqiang Xi and Xiaohan Li}
\thanks{Wenshuo Wang is a Ph.D. candidate in Mechanical Engineering, Beijing
Institute of Technology, Beijing, China. Now he is studying in the
Vehicle Dynamics and Control Lab, University of California at Berkeley.
         (email:wwsbit@gmail.com,wwsvdc2015@berkeley.edu)}
\thanks{Junqiang Xi is a Professor in Mechanical Engineering, Beijing Institute
of Technology, Beijing, China (email:xijunqiang@bit.edu.cn)}
\thanks{Xiaohan Li is a Ph.D. candidate in Faculty of Mechanical Engineering and
Transport Systems, Chair of Human-Machine Systems, Technische Universit\"{a}t
Berlin, Berlin, D-10587, Germany.
        (email:xli@mms.tu-berlin.de)}
}


\maketitle

\begin{abstract}
Driving styles have a great influence on vehicle fuel economy, active safety, and drivability. To recognize driving styles of path-tracking behaviors for different divers, a statistical pattern-recognition method is developed to deal with the uncertainty of driving styles or characteristics based on probability density estimation. First, to describe driver path-tracking styles, vehicle speed and throttle opening are selected as the discriminative parameters, and a conditional kernel density function of vehicle speed and throttle opening is built, respectively, to describe the uncertainty and probability of two representative driving styles, e.g., aggressive and normal. Meanwhile, a posterior probability of each element in feature vector is obtained using full Bayesian theory. Second, a Euclidean distance method is involved to decide to which class the driver should be subject instead of calculating the complex covariance between every two elements of feature vectors. By comparing the Euclidean distance between every elements in feature vector, driving styles are classified into seven levels ranging from low normal to high aggressive. Subsequently, to show benefits of the proposed pattern-recognition method, a cross-validated method is used, compared with a fuzzy logic-based pattern-recognition method. The experiment results show that the proposed statistical pattern-recognition method for driving styles based on kernel density estimation is more efficient and stable than the fuzzy logic-based method.
\end{abstract}


\IEEEpeerreviewmaketitle

\section{Introduction}

Understanding of driving styles plays a crucial role as a supervision and control part in the ``driver-vehicle-environment'' system. Driving behavior and/or driving styles (e.g., reckless and careless driving style, anxious driving style, angry and hostile driving style, patient and careful driving style, and dissociating driving style are defined by \cite{Taub04}) have a big influence on road safety \cite{sagb15}, vehicle performance such as fuel economy \cite{yu12}, comfort\cite{wws15}, etc.. For example, in terms of driving skills, an experienced driver usually drives in an economy way, while a new driver drives in a fuel-consuming way in the same driving environment. For driving styles, an aggressive driver usually prefers to vehicle dirvability performance, inversely, a normal (typical) driver usually prefers to the comfort and fuel economy. Therefore, it is important to recognize  and classify the driving styles to offer a feedback information to the vehicle control system, allowing the intelligent and advanced vehicle control system not only to meet individual driver's needs in time but also to guarantee the safety of surrounding vehicles and the host vehicle. The essential work to achieve this goal is to recognize driver's driving styles or driving characteristics and integrate them into vehicle control systems, which can allow the vehicle to adapt the individual driver. 

Recognizing driving behavior or driving styles is a challenging work since feature parameters will be various for different driving behaviors and driving environments. Fortunately, many related solid works have been done about applications of driver model \cite{Ploc07}, recognition of driver behavior or characteristics \cite{Wang14}, and the human behavior-recognition algorithms \cite{Cand10}, recognition of driver distraction \cite{wangshouyi15}\cite{dong11}.  For recognition methods of driving characteristics, they can be roughly classified into two categories: {\it direct method} (data-based method) and {\it indirect method} (model-based method). These two methods are discussed as follows.

For {\it indirect method}, it requires to establish a driver model that can describe driver's basic driving characteristics such as lane-keeping, lane-changing, and obstacle avoidance, etc. Then, identify or extract driving characteristics based on the parameter estimation of the proposed driver model. Because of the non-linearity and uncertainty of driving behaviours, it is difficult to precisely identify and extract driver characteristics in time. Fortunately, many stochastic process theories can be applied to the recognition of driving behaviours. Hidden Markov Model (HMM), as a simple dynamic Bayesian network, can identify the underlying relationship between observations and states, thus it can be widely utilized to model and predict driver's states \cite{tade14} and driving behaviours \cite{gade14}\cite{mitr05}. In \cite{gade14}, the driver-vehicle system was treated as a hybrid-state system, and  the HMM was used to estimate driver's decisions when driving near intersections. Further, to model and characterise the driving behaviour, the autoregressive with exogenous (ARX) model was widely extended and applied. In \cite{sund13}, a probabilistic ARX model was utilized to predict driver behaviors and classify the driver's driving style, i.e., aggressive driving and normal driving. To mimic and model the uncertainty of driver's behaviour, a stochastic switched-ARX (SS-ARX) model was developed and adopted by Akita et al. \cite{akita07} and Sekizawa et al. \cite{seki07}.

In terms of {\it direct method}, the basic process of the direct method is to directly analyse the driving data (e.g., vehicle speed, steering wheel angle, throttle opening, etc.) using pattern-recognition or data-analysis method without establishing relevant driver models. For recognition of driving skills, Zhang et al. \cite{zhang10} proposed a direct pattern-recognition approach based on three recognition methods, i.e., multilayer perception artificial neural networks (MLP-ANNs), decision tree, and support vector machines (SVMs). The coefficients of discrete Fourier transform (DFT) of steering wheel angles were treated as the discriminant features. In \cite{higg15}, relationships between driver state and driver's actions were investigated using the cluster method with eight state-action variables. For different driving patterns of drivers, the state-action cluster were different, thus segmenting driver into different patterns. 
Though the above mentioned works have made a great a progress in recognition of driver behavior or driving styles/skills/intentions, lots of challenging tasks haven't been overcome, such as the uncertainty of driving styles and preferences that are affected not only by driver's psychological and physical states, but also by driving environments (including the traffic condition, weather, road condition) and vehicle conditions. According to the above discussion, in terms of recognition of driving styles, there are two key issues: 
\begin{itemize}
\item {\it Selection of training data sets}. It is difficult to select a pair of data sets that can represent or define all the aggressive (or normal) driver, though some {\it rule-based strategies} can roughly classify most drivers into different categories. One reason is that the collection driving data of an aggressive driver may not always show a aggressive driving behavior. Another reason is that the aggressive threshold value will be different for individuals.
\item {\it Uncertainty of driving behaviors/styles}. The uncertainty of driver behavior or driving characteristics may be affected by the disturbance coming from driving environments, driver physical or psychological factors will lead different driving styles even for the same driver at different time and/or driving environments. Though the uncertainty of driver behavior and driving characteristics have been applied to vehicle system control\cite{angk11}\cite{angki11}\cite{bichi10}, it is difficult to make a decision for the category (e.g., aggressive or normal) to which the driver is subject when the uncertain factors are involved \cite{schu12}.
\end{itemize}

For the proposed issues, a statistical method is used to describe the uncertainty of driving styles from the point of probability.
In this work, the {direct method} is applied and driving styles of path-tracking attract our attentions, i.e., {\it aggressive driving styles} and {\it normal} (typical) {\it driving styles}. To deal with the uncertainty of driving styles, the driving data are preprocessed using the stochastic (probability) approach. First, a conditional distribution function -- kernel density function -- is involved to describe the uncertainty of driving styles based on the training data, which can be treated as feature parameters of driving styles. Second, according to the kernel density function and full Bayesian theory, the posterior probability of any pair of input data can be calculated against every category. And then, to make a decision to which category the driver are subject and avoid to calculate the complex variance of each pair elements in feature vector, the Euclidean distance is introduced. Last, to shown benefits of the proposed pattern-recognition method using kernel density estimation, the cross-validation experiment is conducted by comparing with a fuzzy logic recognition method.

Following the overview in the first section of this paper, Section \Rmnum{2} discusses the way of selection of feature parameters. Section \Rmnum{3} presents the fuzzy logic recognition algorithm and the proposed method, including kernel density and Euclidean distance. Section \Rmnum{4} presents the data collection and experiment design in driving simulator. Then the experiment results  and analysis are shown in Section \Rmnum{5}. Last, the discussion and the conclusion about this work are shown in Section \Rmnum{6}.

\section{Feature Parameters Selection}
The goal of feature parameter selection is to allow pattern vectors belonging to different categories to occupy compact and disjoint regions as much as possible in a $ d $-dimensional feature space \cite{Jain00}\cite{Dunne07}. Generally, the data using for pattern recognition of driving styles could be classified into three categories: 1) {\it driver-dependent}, which includes the physical signal and physiological signal that are directly related to the human driver. In terms of physical signal, for example, the steering angle, brake force, and throttle opening signals, gesture signal, eyes related signal are used for recognition of aggressive or normal driver\cite{Quin12}\cite{Higg13}, driver fatigue or drowsiness \cite{dong11}, driver's intentions\cite{cheng06}\cite{kuma13}\cite{kasp12} etc.. For physiological signal, rate of hear beat, EGG, or EMG signal is usually used to recognize driver emotions or tension levels \cite{heal05}, etc.; 2) {\it vehicle-dependent}, which includes vehicle speed, acceleration, yaw angle, are used for recognition of driving skills levels or driver behaviors \cite{Miya07}\cite{Ma07}\cite{Quin12}\cite{Higg13}.; and 3) {\it driving environment-dependent}, which includes road profile, road-tire coefficients, surrounding vehicles, traffic flow, etc.. In this work, the driver path-following (path-tracking) behavior on the curve road is focused on and the vehicle-dependent signals are selected for recognition of driving styles. To characterize driver behavior with different driving tasks, the feature parameter should be selected first. For different driving task, the parameter selection will be different(see Table \ref{Table1}). 

\begin{table}[h]
\begin{center}
\caption{Feature parameters selection and pattern-recognition method associated with relevant driving characteristics\label{Table1}}
\begin{tabular}{p{0.9in}|p{1.1in}|p{0.9in}}
\hline
\hline
Driving task & Feature parameters & Method \\
\hline
Car-following \cite{higg15}\cite{Miya07}\cite{Ma07} & \tabincell{l}{--Distances between cars; \\--Vehicle speed; \\--Vehicle Position} & \tabincell{l}{--GMM; \\--FC}\\
\hline
Driving skill \cite{zhang10} & --Steering angle                                           & \tabincell{l}{--SVM; \\ --DT; \\ --MLP-NN} \\
\hline 
Curve negotiating (or curve path-tracking) & --Vehicle speed                 & \tabincell{c}{--MPC} \\
\hline
Driving styles \cite{sund13} \cite{Quin12}-\cite{kasp12} & \tabincell{l}{--Acceleration; \\--Yaw rate; \\--Lateral displacement; \\--Vehicle Speed; \\--Steering angle;\\--Physical signal;\\--Physiological signal} & \tabincell{l}{--P-ARX; \\ --NN; \\--FL; \\--HMM; \\--SVM; \\--OOBNs or BNs; \\--Bayesian filter} \\
\hline
\hline
\multicolumn{3}{p{3in}}{\textbf{NOTES}: GMM--Gaussian Mixture Model; FC -- Fuzzy Clustering; SVM -- Support Vector Machine; DT -- Decision Tree; MLP-NN -- Multilayer Perception-Neural Network; MPC -- Model Predictive Control; P-ARX -- Probabilistic Autoregressive Exogenous; NN -- Neural Network; FL -- Fuzzy Logic; HMM -- Hidden Markov Model; OOBNs -- Objected-oriented Bayesian Networks} \\
\hline
\hline
\end{tabular}
\end{center}
\end{table}

\begin{figure*}[thpb]
      \centering
     \includegraphics[width=1\linewidth]{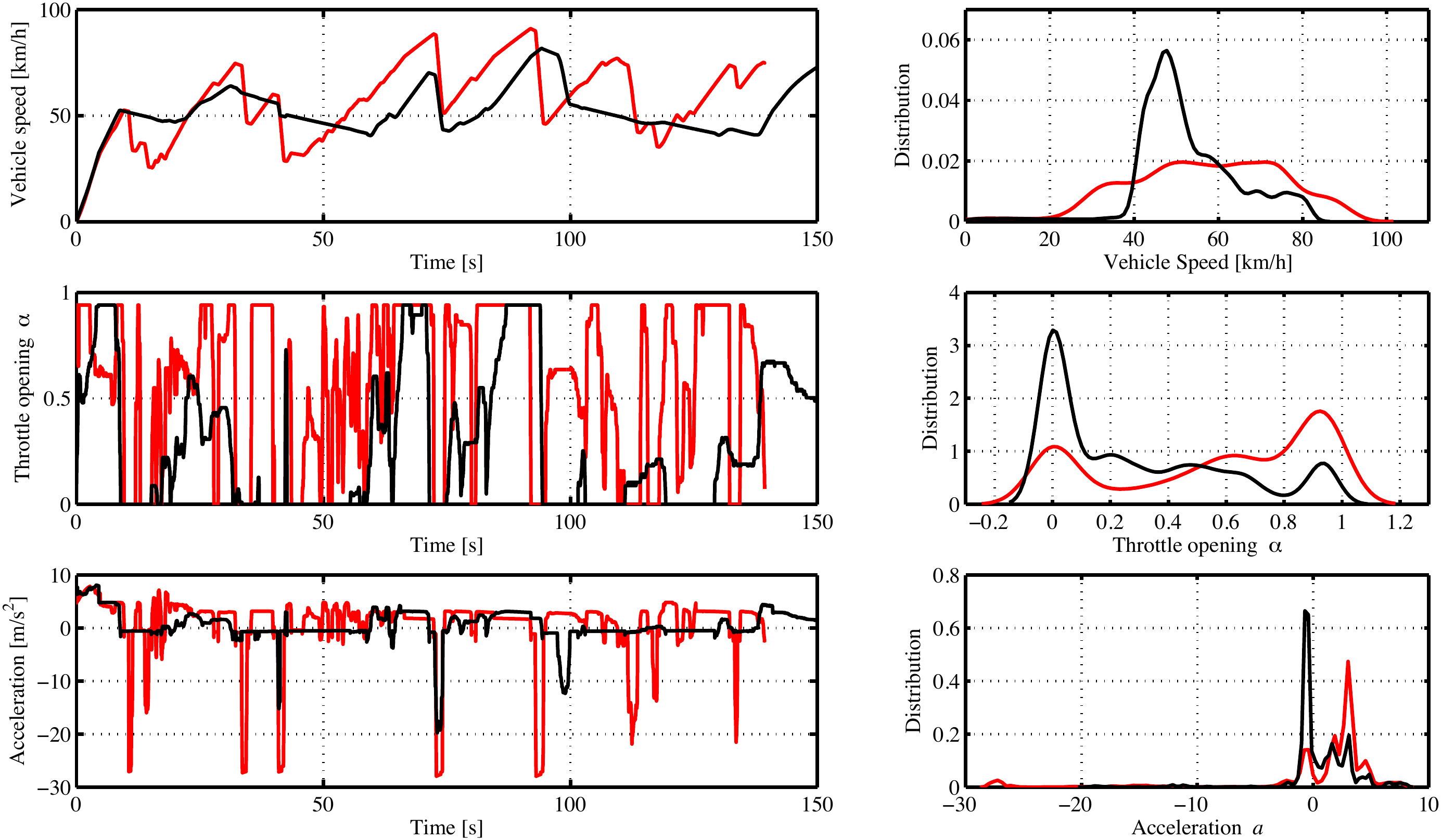}
      \caption{Driving data (Left column) and its distribution (Right column) for two kinds of driving styles. Red line  ({\color{red}{--}}) represents the aggressive driver and the black line ({\color{black}{--}}) represents the (normal) typical driver.}
      \label{figure2}
   \end{figure*}

To select a feature parameter that can describe driving styles when taking a curve-negotiation task, the distribution analysis of parameters is conducted. Fig. \ref{figure2} shows the basic driving data and corresponding distributions of two driver with different driving styles. To select a series of feature parameters, assumptions are made as follows:

\begin{itemize}
\item {\it Statistical characteristic invariance}: Given a constant driving environment, the vehicle speed or throttle opening driver selected are relatively variable, but its statistical property, such as distribution property, could  be treated as invariance to some degree. For example, an aggressive driver prefers a high vehicle speed than a normal driver, and the vehicle speed prefers to fall a constant interval $ [\underline{v}_{s} \pm  \varepsilon,\overline{v}_{s} \mp \varepsilon] $, $ \varepsilon $ is a small positive value. Table \ref{Table2} shows two drivers with distinct driving styles (aggressive and normal) and the statistical results indicate that the invariance of statistical property is nearly constant for one selected driver.
\item {\it Maximum discrimination}: The selected feature parameters, to some degree, should maximize the discrimination between different driving styles. From the statistical results in Fig. \ref{figure2}, it is obvious that the vehicle speed driver selects and throttle opening driver controls are suitable to be feature parameters.
\end{itemize}
Therefore, based on these assumptions and Fig. \ref{figure2}, vehicle speed ($ v $) and throttle opening ($ \alpha $) are selected as feature parameters ($ \boldsymbol{x}=(v,\alpha) $) to describe driver path-tracking behavior in different kinds of styles and discussed as followed.

\begin{table}[h]
\caption{Means and variances of vehicle speed and throttle opening for two different drivers\label{Table2}}
\begin{center}
\begin{tabular}{l|l|l|l}
\hline
\hline
\multicolumn{2}{c|}{Driver 1 (Aggressive)}  & \multicolumn{2}{c}{Driver 2 (Normal)}  \\
\hline
Mean(Var)\_Sp & Mean(Var)\_Th &  Mean(Var)\_Sp & Mean(Var)\_Th \\
\hline
   56.849(317.551)  &  0.566(0.131)  &     52.492(152.317)  &  0.285(0.099)\\
   61.334(256.959)  &  0.610(0.133)  &     48.530(173.298)  &  0.259(0.063)\\
   61.928(250.443)  &  0.645(0.136)  &     52.925(129.773)  &  0.235(0.071)\\
   61.336(273.581)  &  0.625(0.127)  &     50.565(137.977)  &  0.238(0.059)\\ 
   64.451(307.289)  &  0.599(0.146)  &     50.666(106.973)  &  0.195(0.056)\\
   64.241(301.918)  &  0.688(0.120)  &     50.531(117.186)  &  0.213(0.064)\\
   63.146(296.334)  &  0.612(0.142)  &     49.487(154.639)  &  0.284(0.057)\\
   61.932(263.429)  &  0.649(0.142)  &     46.344(115.193)  &  0.156(0.030)\\
   64.147(287.705)  &  0.608(0.126)  &     48.247(95.392)   &  0.153(0.037)\\
\hline
\hline
\end{tabular}
\end{center}
\end{table}

\subsection{Vehicle Speed}
For driving styles when tracking a given road, the vehicle speed is one of parameters that can directly show and characterize driver behavior and driving preferences \cite{Rich12}, such as aggressive or normal styles, shown as Fig. \ref{figure2} and Table \ref{Table2}.
For example, in Fig. \ref{figure2} and Table \ref{Table2}, it is obvious that the aggressive prefers to vehicle speeds $ v_{agg} \in \{ [20,40] \cup [60,100] \} $ km/h, while the normal driver prefers to vehicle speeds $ v_{norm} \in [40,60]$ km/h.

\subsection{Throttle Opening}
As one of the direct control parameter by the human driver, throttle openings can directly reflect driver's preferences or driving styles. Besides, the distributions of throttle openings for different drivers (shown in Fig. \ref{figure2} and Table \ref{Table2}) are more distinct than accelerations. In Table \ref{Table2}, Mean\_$ _{(\cdot)} $ and Var\_$ _{(\cdot)} $ represent the mean and variance value of $ (\cdot) $, $ (\cdot)\in \{ SP=\mbox{speed}, Th=\mbox{throttle opening} \} $. For example, Mean\_$ _{SP} $ represents the mean value of vehicle speed.

Therefore, the vehicle speed $ v $ and throttle opening $ \alpha $ are selected as the feature parameters $ \boldsymbol{x}= $($ v,\alpha $) to represent driving styles. And then, based on the selected feature parameter $ \boldsymbol{x} $, a model $ f $ should be trained using the training data to recognize driving styles $ s \in \mathcal{S}=\{ s|s=-3,-2,-1,0,1,2,3 \} $, i.e., $ f:\boldsymbol{x} \rightarrow s $. Here, the element of set $\{ -3,-2,-1,0,1,2,3 \} $ represents the aggressive or normal level. A  lager value of $ s $ means that a more aggressive driving styles. For example, $ s = -3 $ represents a lowest normal driver and $ s = 3 $ represents a high aggressive driving style.

\section{Method}

In this section, pattern-recognition approaches of driving styles are discussed. First, a fuzzy-logic pattern-recognition approach is presented in the first subsection to make a comparison with the proposed recognition method. Second, a pattern-recognition is developed using kernel density estimation and Euclidean distance from the point of statistical distribution. The kernel density estimation method is introduced to estimate driver's preference based on training data. Subsequently, to identify the level of driving styles, the Euclidean distance is involved. 

\subsection{Fuzzy Logic}

The recognition of driving styles is a {\it vague} concept that can not precisely divide drivers into the defined categories, such as aggressive or normal, as for different drivers the aggressive or normal scales of driving behavior will be different. Driver behavior, which can be treated as a {\it natural language}(driving behavior language), is vague. Our perception of the real driving styles is pervaded by concepts which do not have sharply defined boundaries. Therefore, a mathematical tool called  {\it fuzzy logic}(FL) is widely developed and introduced for recognition driver manoeuvre \cite{Huln10}, driving profile \cite{Wahab09} or driving styles and dealing with the uncertainty of driving styles, which provides a technique to deal with imprecision and information granularity.

In this work, the fuzzy inference system(FIS) based on Mamdani rule are defined as a fuzzy recognition system with two inputs (i.e., vehicle speed and throttle opening) and one output(i.e., level of driving styles). The definition of membership function are defined based on an expert driver knowledge.

\begin{figure}[thpb]
\centering
\includegraphics[width=1\linewidth]{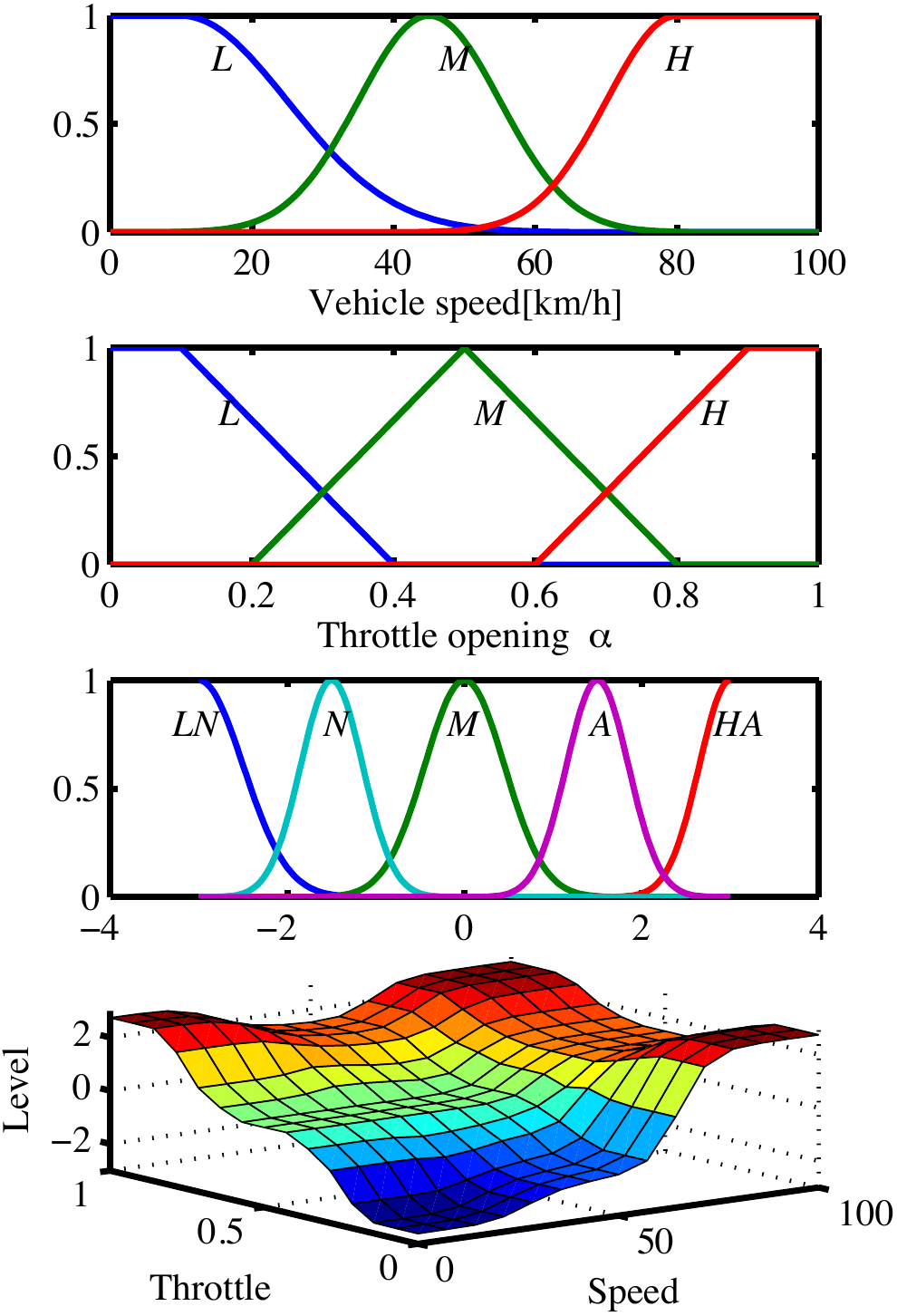}
\caption{Membership function for the inputs and output of FIS. Top1: membership functions of vehicle speed. Top2: membership functions of throttle opening. Top3: membership functions of output. Last: whole process mapping of FIS.}
\label{MembershipFigure}
\end{figure}

Corresponding fuzzy values of the first input, vehicle speed($ v $), are defined to be {\it lower} ($ L $), {\it middle} ($ M $) and {\it high} ($ H $). The fuzzy values of second input, throttle opening($ \alpha \in [0,1] $),  are defined to be {\it lower} ($ L $), {\it middle} ($ M $) and {\it high} ($ H $). The fuzzy values of output, level of driving styles, are defined to be {\it lower normal} ($ LN $), {\it normal} ($ N $), {\it middle} ($ M $), {\it aggressive} ($ A $), and {\it high aggressive} ($ HA $). Here we code the output sets $ LN $ and $ HA $ as $ -3 $ and $ 3 $.  All membership functions are shown in Fig. \ref{MembershipFigure} and the fuzzy rules are defined in Table \ref{StyleRules}.

\begin{table}[h]
\caption{Fuzzy rules for definition of driving styles\label{StyleRules}}
\begin{center}
\begin{tabular}{cccccc}
\hline
\hline
Index & Input1 & Operator & Input2 & Weight & output  \\
\hline
1     &  L     & and  & L      &   1     &  LN              \\
2     &  L     & and  & M      &   1     &  M              \\
3     &  L     & and  & H      &   1     &  HA              \\
4     &  M     & and  & L      &   1     &  N              \\
5     &  M     & and  & M      &   1     &  M              \\
6     &  M     & and  & H      &   1     &  A              \\
7     &  H     & and  & L      &   1     &  HA              \\
8     &  H     & and  & M      &   1     &  A              \\
9     &  H     & and  & H      &   1     &  HA              \\
\hline
\hline
\end{tabular}
\end{center}
\end{table}

\begin{figure*}[thpb]
      \centering
\includegraphics[scale = 0.6]{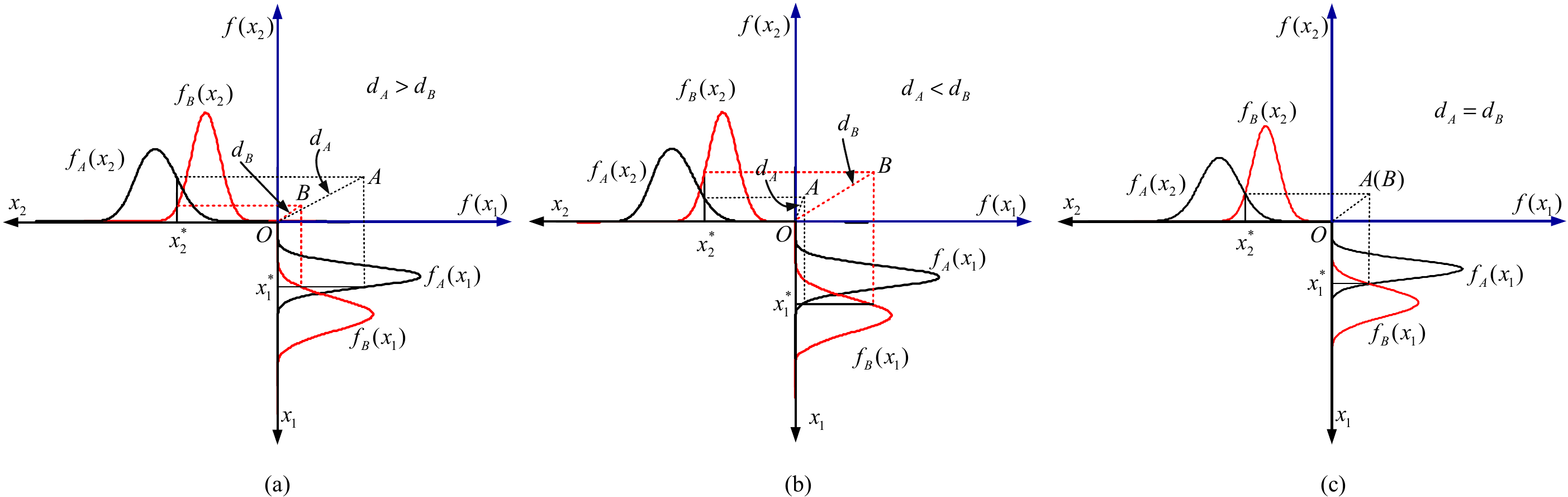}
      \caption{Schematic Diagram of the proposed pattern-recognition method of driving styles using kernel density estimation and Euclidean distance. Here, $ f_{(\cdot)}(x_{k})=P(\cdot|x_{k}) $ represents the posterior probability, i.e., the probability that $ x_{k} $ belongs class $ (\cdot) $. (a)Decide input data $ (x_{1}^{\ast}, x_{2}^{\ast}) \in A $; (b) decide input data $ (x_{1}^{\ast}, x_{2}^{\ast}) \in B $; and (c) decide input data $ (x_{1}^{\ast}, x_{2}^{\ast}) \in $ fuzzy class $ M $.}
      \label{figure1}
\end{figure*}

\subsection{Proposed method}
\subsubsection{Kernel Density Estimation}
Kernel density estimation, as an unsupervised learning method, can estimate a probability density at a point $ x_{0} $ given a random sample $ x_{1}, x_{2},\cdots,x_{N} $ from a probability density $ f(x) $. For two classes of 1-Dimension data sequences $ X_{1}=\{ x_{1}^{1},\cdots,x_{i}^{1},\cdots,x_{n}^{1} \} \in \mathcal{C}_{1} $  and  $ X_{2}=\{ x_{1}^{2},\cdots,x_{j}^{2},\cdots,x_{m}^{2} \} \in \mathcal{C}_{2} $, where $ x_{i}^{1},x_{j}^{2}\in \mathbb{R}^{1\times1} $, $ X_{1} \in \mathbb{R}^{n} $, and $ X_{2} \in \mathbb{R}^{m} $, we can get two {\it class-conditional probability density} functions $ f(x|\mathcal{C}_{1}) $ and $ f(x|\mathcal{C}_{2}) $\cite{duda00}, also called {\it likelihood}. In this work, the Gaussian kernel density at point $ x_{0} $ is used to calculate the probability density $ f(x|X) $.

\begin{equation}\label{Eq1}
\begin{split}
p(\boldsymbol{x}|\mathcal{C}_{k}) &  =f(x_{0}|X)  \\
             & =\frac{1}{N\lambda}\sum_{i=1}^{N}{K_{\lambda}(x_{0},x_{i})} \\
             & =\frac{1}{N(2 \lambda^2 \pi)^{\frac{d}{2}}} \sum_{i=1}^{N}{exp\left\lbrace -\frac{1}{2} (\frac{\| x_{i} - x_{0} \|}{\lambda})^2  \right\rbrace } \\
\end{split}
\end{equation}
where $ K_{\lambda} $ is the Gaussian kernel.

\subsubsection{Bayesian Theory and Bayesian Decision}
Suppose that the prior probabilities $ P(\mathcal{C}_{k}) $ and the {\it conditional-probabilities density} $ p(\boldsymbol{x}|\mathcal{C}_{k}) $, also called {\it posterior probability}, are know for  the number of class $ k = 1,2,\cdots $. Based on {Bayes formula}:

\begin{equation}\label{Eq2}
\begin{split}
P(\mathcal{C}_{k}|\boldsymbol{x}) & =\frac{p(\boldsymbol{x}|\mathcal{C}_{k})P(\mathcal{C}_{k})}{p(\boldsymbol{x})} \\
p(\boldsymbol{x})  &   = \sum_{k=1}^{l}{p(\boldsymbol{x}|\mathcal{C}_{k})p(\mathcal{C}_{k})}\\
\end{split}
\end{equation}
And then, the {\it posterior probability} given random input $ \boldsymbol{x} $ can be estimated by Equation \eqref{Eq2}. Under \eqref{Eq2}, decisions about $ \boldsymbol{x} $ can be made:

\begin{equation}\label{Eq3}
\mbox{Decide} \ \mathcal{C}_{k} \ \mbox{if} \ P(\mathcal{C}_{k}|\boldsymbol{x}) > P(\mathcal{C}_{\setminus k}|\boldsymbol{x}) 
\end{equation}
where $ P(\mathcal{C}_{\setminus k}|\boldsymbol{x}) $ represents all the left categories except for the $ k $th category.

The key to calculate \eqref{Eq2} and decide \eqref{Eq3} is the conditional probability density $ p(\boldsymbol{x}|\mathcal{C}_{k}) $. However, for a higher-dimension feature vector in a higher-dimension feature space, the calculation for covariances of every pair of dependent components in feature vectors will be more difficult. Therefore, instead of calculating the complicated covariances, a Euclidean distance-based making-decision method is proposed in the following subsection.

\subsubsection{Decision Making Using Euclidean Distance}
{\it Bayesian decision} can be easily used in 1-Dimension case, but for a $ d $-dimension case ($ d \geq 2 $) and the elements in  {\it feature vector} $ \boldsymbol{x} $ are highly dependent, it is difficult to calculate the {\it conditional-probability} $ p( \boldsymbol{x}|\mathcal{C}_{k} ) $. To overcome the issue, the {\it Euclidean Distance} is involved to deal with the decision-making issues, here, to determine to which category the driver is subject and measure the levels of driving aggressive/normal type. 

Taking two classes (class $ A $ and class $ B $) with 2-Dimension($ d=2 $) case for example (Fig. \ref{figure1}), the  posterior probability that elements in feature vector $ \boldsymbol{x}=(x_{1},x_{2}) $ are belonged to class $ A $ and $ B $ is defined as $ f_{A}(x_{l})$, $f_{B}(x_{l}) $ for $ l=1,2 $, respectively. Here, $ f_{A}(x_{l})=P(A|x_{l})$, $f_{B}(x_{l})=P (B|x_{l})$ . Given random input $ \boldsymbol{x}^{\ast}=(x_{1}^{\ast},x_{2}^{\ast}) $, the relevant posterior probability $ f_{A}(x_{l}^{\ast}), f_{B}(x_{l}^{\ast}) $ can be calculated by \eqref{Eq1} and \eqref{Eq2} for $ l=1,2 $. Given any input and project their corresponding posterior probability into the first quadrant, getting $ A=(f_{A}(x_{1}^{\ast}),f_{A}(x_{2}^{\ast})) $ and $ B=(f_{B}(x_{1}^{\ast}),f_{B}(x_{2}^{\ast})) $, and the Euclidean distance is defined as

\begin{equation}\label{Eq4}
\begin{split}
d_{A} := & \parallel f_{A}(x_{1}^{\ast})^{2} + f_{A}(x_{2}^{\ast})^{2} \parallel^{\frac{1}{2}} \\
d_{B} := & \parallel f_{B}(x_{1}^{\ast})^{2} + f_{B}(x_{2}^{\ast})^{2} \parallel^{\frac{1}{2}} \\
\end{split}
\end{equation}

The joint density function $ p((x_{1}^{\ast},\cdots,x_{d}^{\ast})|\mathcal{C}_{k}) $ of $ d $-dimension feature vectors ($ d \geq 2 $) is decoupled into several simple densities of 1-dimension feature scalar and the Bayesian decision is transformed to the Euclidean distance-based decision. The decision-making rules based on Euclidean distance are defined as follows:

\begin{itemize}
\item Decide class $ A $ if $ (d_{A} > d_{B}) \wedge (| d_{A} - d_{B} | > \epsilon) $ (Fig. \ref{figure1}(a))
\item Decide class $ B $ if $ (d_{A} < d_{B}) \wedge (| d_{A} - d_{B} | > \epsilon) $ (Fig. \ref{figure1}(b))
\item Decide class $ M $ if $ | d_{A} - d_{B} | \leq \epsilon $ (Fig. \ref{figure1}(c))
\end{itemize}
where $ M $ is the fuzzy class between class $ A $ and $ B $, $ \epsilon $ is a positive threshold value, $ \epsilon \in \mathbb{R}^{+} $. 

We should note that, when $ \boldsymbol{x} $ is in a $ d $-dimensional Euclidean space $ \mathbb{R}^{d} $ for $ d=3 $, the Euclidean distance is the radius of a sphere. Therefore, the expanded Euclidean distance can be calculated by:

\begin{equation}\label{Eq5}
d_{\mathcal{C}_{k}}(x_{i}^{\ast}) := \parallel \sum_{i=1}^{d} f_{\mathcal{C}_{k}}(x_{i}^{\ast})^{2} \parallel^{\frac{1}{2}}, k=1,2,\cdots
\end{equation}
for any input $ \boldsymbol{x}^{\ast}=\left( x_{1}^{\ast},\cdots, x_{i}^{\ast}, \cdots, x_{d}^{\ast} \right)  $.

\subsubsection{Classification Algorithm}
Based on the above description, a classification method based on conditional-kernel density $ f_{\mathcal{C}_{k}}(x) $ and Euclidean distance $ d_{\mathcal{C}_{k}} $ is developed. To represent different level of driving types simply, a number set is involved as $ \mathcal{S}=\{-3,-2,-1,0,1,2,3 \} $. A lager value of number indicates a more aggressive driving styles. The classification algorithm is shown in Table \ref{Table3}. In Table \ref{Table3}, the threshold value $ ( \underline{\epsilon},\overline{\epsilon} ) $ and $ (\underline{\epsilon}^{\star},\overline{\epsilon}^{\star}) $ are selected as Table \ref{Table4}. For training step $ 3 $, the prior probability $ P(\boldsymbol{x}) $ is set to $ 1/k $, $ k $ is the number of classification
for training data. In this work, $ k=2 $, two typical driving styles of training data, i.e., aggressive and normal, are considered.

\begin{table}[h]
\caption{Algorithm of the proposed recognition approach for driving styles\label{Table3}}
\begin{center}
\begin{tabular}{p{0.04in} p{2.7in}}
\hline
\hline
\textbf{Training} & \\
1: & Input training data sequence $ \mathcal{X}^{k}=\{ \boldsymbol{x}_{i}^{k} \} $ for $ k =1,2$, $ i=1,\cdots,n $,  $ \boldsymbol{x}_{i}^{k}=\{x_{i,l}^{k}\} $, $ l=1,\cdots,d $ \\
2: & Get the conditional-kernel density estimation function $ f(x_{i,l}^{k}|{\mathcal{C}_{k}}) $ under Equation \eqref{Eq1} for each single component in feature vectors\\
\textbf{Testing} & \\
1: & Input new data $ \boldsymbol{x}_{i}^{\ast}= \{x_{i,l}^{\ast} \} $, $ i=1,2,3,\cdots $ \\
2: & \underline{\textbf{for}} $ i,l $ \\
3: & \quad Get $ f({x_{i,l}^{\ast}|\mathcal{C}_{k}}) $ from  $ f(x_{i,l}^{k}|{\mathcal{C}_{k}}) $ for $ k=1,2 $\\
4: & \quad Get $ P(\mathcal{C}_{k}|x_{i,l}^{k}) $ under Equation \eqref{Eq2} and set $ f_{\mathcal{C}_{k}}(x_{i,l}^{\ast}) = P(\mathcal{C}_{k}|x_{i,l}^{k}) $ for $ k=1,2 $\\
5: & \quad Get $ d_{\mathcal{C}_{k}(x_{i,l}^{\ast})} $ $ \Leftarrow $ Equation \eqref{Eq5} \\
6: & \quad \quad \underline{\textbf{if}} $ d_{\mathcal{C}_{k}(x_{i,l}^{\ast})} > d_{\mathcal{C}_{\setminus k}(x_{i,l}^{\ast})} $ \\
7: & \quad \quad \ \ \ \underline{\it \textbf{if}} $ d_{\mathcal{C}_{k}(x_{i,l}^{\ast})} - d_{\mathcal{C}_{\setminus k}(x_{i,l}^{\ast})}  \in (\underline{\epsilon},\overline{\epsilon})$ (see Table \ref{Table4})  \\
8: & \quad \quad \ \ \ \ then $ \boldsymbol{x}_{i}^{\ast} \in $ level $ s = \mathcal{S} $ \\
9: & \quad \quad \ \ \ \underline{\it \textbf{end if}} \\
10: & \quad \quad \underline{\textbf{else}} $ d_{\mathcal{C}_{k}(x_{i,l}^{\ast})} \leq d_{\mathcal{C}_{\setminus k}(x_{i,l}^{\ast})} $ \\
11: & \quad \quad \ \ \ \underline{\it \textbf{if}}  $ \parallel d_{\mathcal{C}_{k}(x_{i,l}^{\ast})} - d_{\mathcal{C}_{\setminus k}(x_{i,l}^{\ast})} \parallel \in (\underline{\epsilon}^{\star},\overline{\epsilon}^{\star})$ (see Table \ref{Table4}) \\ 
12: & \quad \quad \ \ \ \ then $ \boldsymbol{x}_{i}^{\ast} \in $ level $ s = \mathcal{S} $ \\
13: & \quad \quad \ \ \ \underline{\it \textbf{end if}} \\
14: & \quad \quad \underline{\textbf{end if}} \\
15: & \underline{\textbf{end for}} \\
16: & Output the classification for sequences data $ \{ \boldsymbol{x}_{i}^{k} \} $ \\
\hline
\hline
\end{tabular}
\end{center}
\end{table}

\begin{table}[h]
\caption{Threshold value of ($ \underline{\epsilon},\overline{\epsilon} $) and ($ \underline{\epsilon}^{\ast},\overline{\epsilon}^{\ast} $)\label{Table4}}
\begin{center}
\begin{tabular}{lc||lc}
\hline
\hline
$ ( \underline{\epsilon},\overline{\epsilon} ) $ & Aggressive style level & $ (\underline{\epsilon}^{\star},\overline{\epsilon}^{\star}) $ & Normal style level \\
\hline
(0.5,-)    & 3 & (0.5,-)    & -3 \\
(0.2,0.5]  & 2 & (0.1,0.5]  & -2 \\
(0.02,0.2] & 1 & (0.02,0.1] & -1 \\
(0,0.02]   & 0$ ^+ $ & [0,0.02]   & 0$ ^- $ \\
\hline
\hline
\end{tabular}
\end{center}
\end{table}

\section{Experiment}
In this section, to show benefits of the proposed recognition approach of driving style, a series of path-tracking tests on the curve road for different participants is conducted in the driving simulator. 

\subsection{Driving Simulator}
\begin{figure}[thpb]
      \centering
      \includegraphics[scale = 0.55]{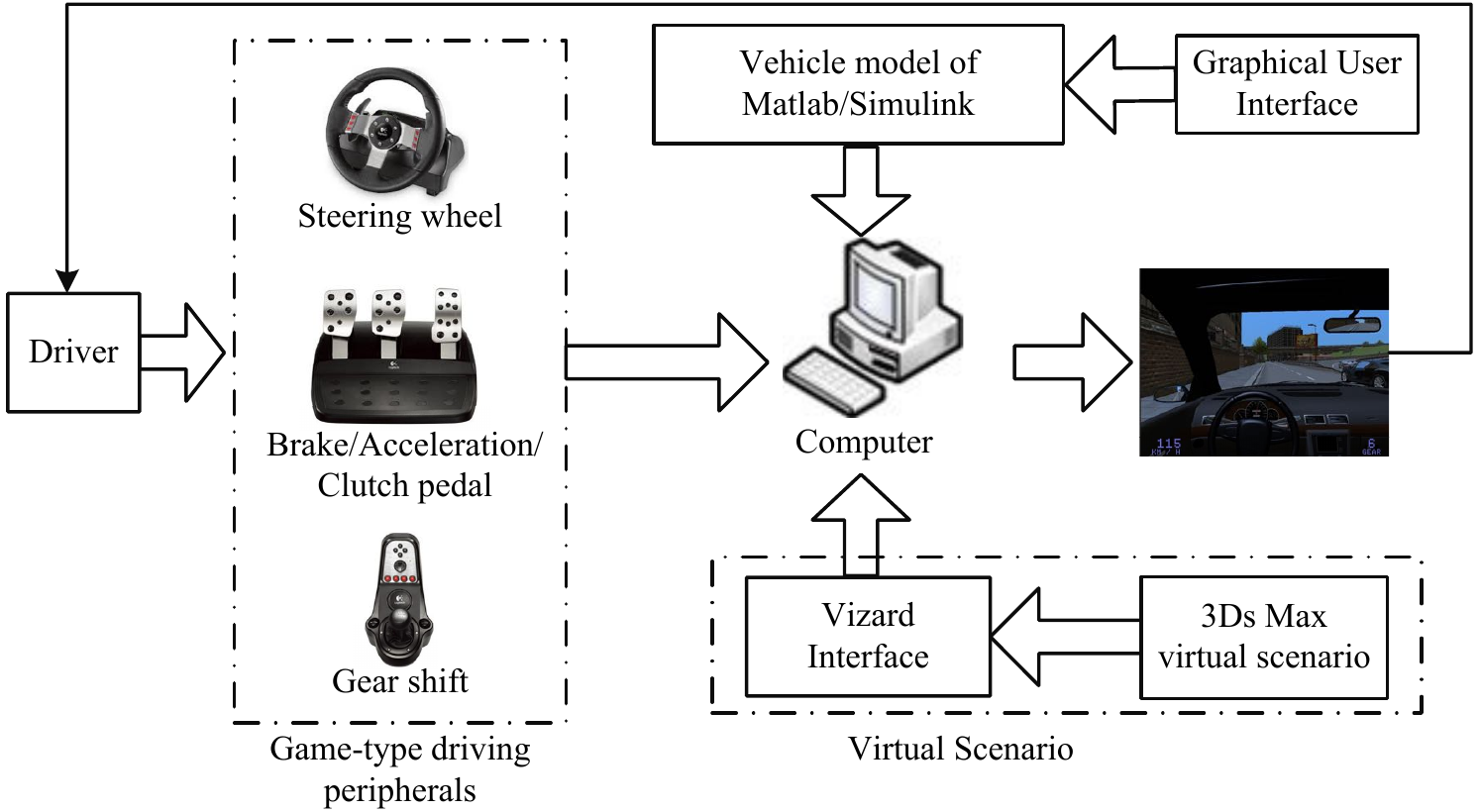}
      \caption{Schematic digram of driving simulator using for data collection \cite{wang15}.}
      \label{DrivingSimulator}
   \end{figure}

All the experiment data are obtained through a driving simulator(See paper \cite{wang15} and Fig. \ref{DrivingSimulator}).  The direct driving data (i.e., driver inputs, including steering angle, throttle opening, braking forces) is input through the game-type driving peripherals. A bicycle-vehicle model is used as the vehicle system. 

\subsection{Road Curve}
\begin{figure}[thpb]
      \centering
\includegraphics[scale = 0.7]{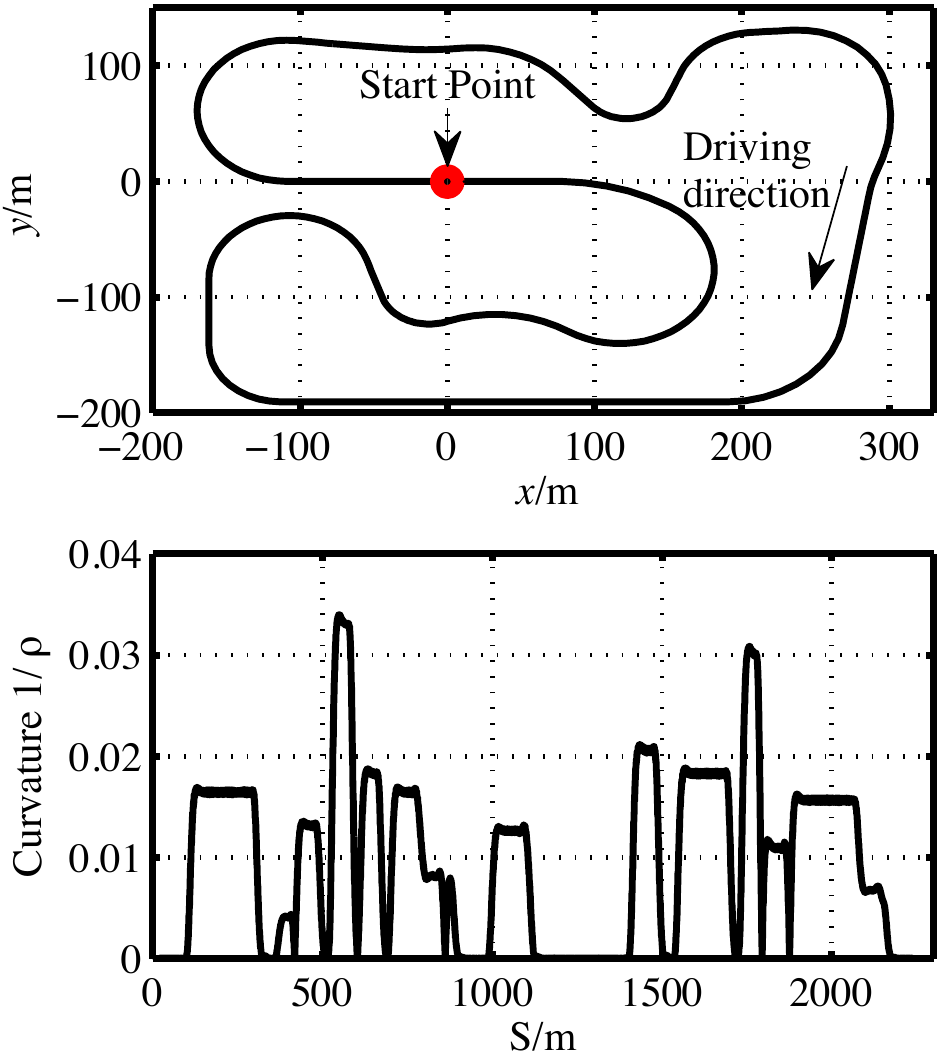}
      \caption{The road profile (Top) and its curvature (Bottom).}
      \label{Roadcurve}
   \end{figure}

In this work, the driver path-tracking in a curve road is focused. The road factors(e.g., road profile) have a big effect on detection of driving style patterns. To design a lifelike driving environment, the road model must have the same scale as road in the real driving environment. To except the effects of natural factors and other disturbs, a special road curve is designed and the natural factors are not taken into consideration. Therefore, the requirements of road model are subject to following criteria: continuity of the path, continuity of the curvature, and differentiability of the set path \cite{kien05}. Therefore, a curve road is designed as in Fig. \ref{Roadcurve}.

\subsection{Test Method}  
All the driving data were collected at a sample frequency at $ 50 $ Hz in the driving simulator, including vehicle speed ($ v $), throttle opening ($ \alpha $), acceleration ($ a $), vehicle position $ (x,y) $, steering angle ($ \delta $), yaw angle ($ \varphi $), etc.  Eighteen people are selected as the participant, nine of them are aggressive drivers and other left half part are normal drivers.    Each participant should be labelled as aggressive or normal before doing a test. During the test, every driver should follow the rules:

\begin{itemize}
\item All participants must be in mentally and physically normal states. 
\item The secondary tasks are forbidden. For example, text a message or answer a phone while driving.
\item Each participant should rest 1 minute before the next run. 
\item Each driver drives a car in their own driving style in the driving simulator. 
\end{itemize}

\section{Recognition Performance Evaluation}
In this section, the experiment results and analysis are discussed to show benefits of the proposed recognition method.

\subsection{Cross-validation}
{\it Cross-validation} (CV) method, as one of most popular evaluation scheme, was used to evaluate the recognition performance of the proposed approach. To do CV, the available training data set is evenly divided into $ q $ parts, called {\it folds}. All folds except random one of folds are used for training recognition model, and the {\it hold-out set} or {\it validation set} is used for assessing the training model.  In this work, driving data sets are evenly divided into nine folds and five folds are used for training and four fold is used for the performance measure of the proposed method and algorithm. CV assessment approach makes sure that the {\it training data sets} are disjoint from the {\it validation sets}.
To evaluate the proposed recognition method, the validation sets are grouped by aggressive and normal drivers to test how well the recognizer may identify them from those provided by the aggressive drivers.

The correction recognition rate (CRR) of driving styles recognizer is defined as:

\begin{equation}\label{Eq6}
CRR_{a} = \frac{Num_{a,a}}{\sum_{\star\in\{ a,n \}}{Num_{a,\star}}}
\end{equation}
for an aggressive driver, and 
\begin{equation}\label{Eq7}
CRR_{n} = \frac{Num_{n,n}}{\sum_{\star\in\{ a,n\}}{Num_{n,\star}}}
\end{equation}
for a normal driver. The first and second subscription of $ Num_{\star,\star} $ are the actual driving styles and the driving styles recognized by the proposed method, respectively. $ a $ and $ n $ represents ``aggressive drivers'' and ``normal drivers'', respectively. Taking $ Num_{a,n} $ for example, it represents the number of runs that are grouped as aggressive drivers and classified as being normal drivers. 

%

\subsection{Results and Analysis}

Fig. \ref{figure3} $ \sim $ Fig. \ref{figure6} show the recognition result for aggressive drivers and normal drivers, respectively, and classify them into different levels using the proposed recognition approach and FL algorithm. The feature analysis, efficiency analysis, and stability analysis are discussed as follow.

\subsubsection{Feature Analysis} 
\begin{figure*}[!t]
      \centering
      \includegraphics[scale = 0.6]{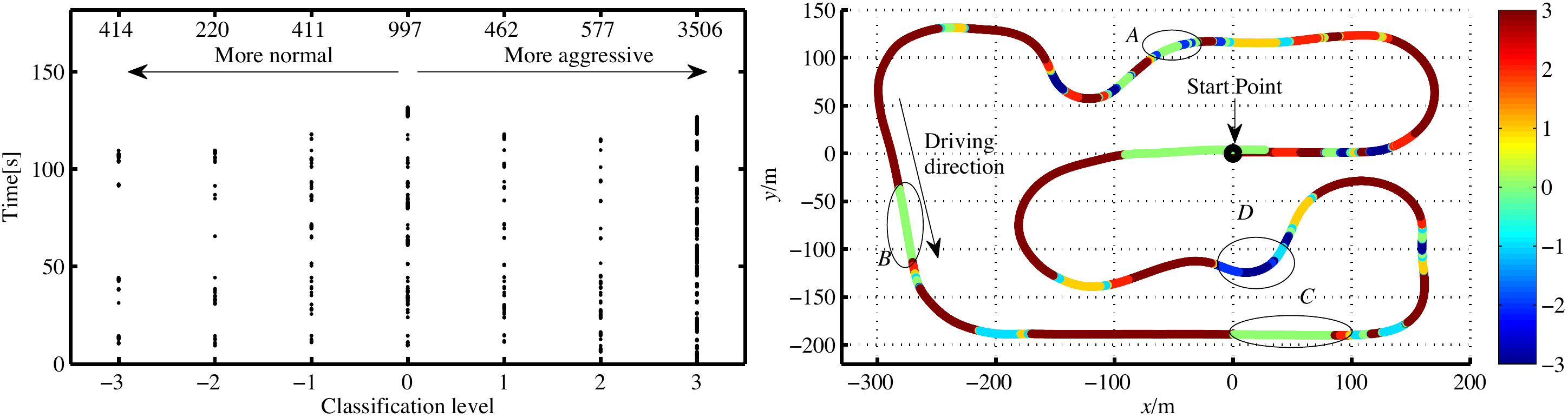}
      \caption{Recognition results for an aggressive driver using the proposed recognition method.(Left: the classification level; Right: the classification result when driving on the curve road.)}
      \label{figure3}
   \end{figure*}

From Fig. \ref{figure3} and \ref{figure5}, it is obvious that the aggressive driver's driving behaviors are mostly labelled as being an aggressive driver by using the proposed recognition algorithm and FL algorithm, only the behaviors at the begin of runs are labelled as being the normal driver. Furthermore, we found that an aggressive driver may show a normal driver behavior before entering a road curve( shown as part $  A $, $ B $, $ C $, $ D $ in Fig. \ref{figure3} and part $  A $, $ B $, $ C $, $ D $, $ E $, in Fig. \ref{figure5}), but after entering the curve road, an aggressive driver will show an aggressive behavior. 

\begin{figure*}[thpb]
      \centering
      \includegraphics[scale = 0.6]{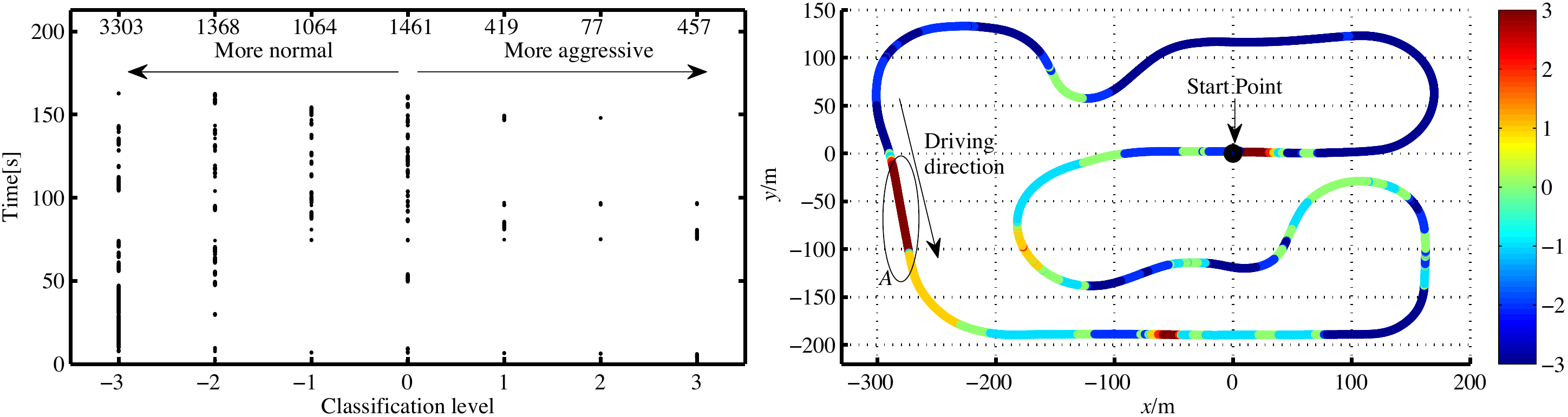}
      \caption{Recognition results for a normal driver using the proposed recognition method. (Left: the classification level; Right: the classification result when driving on the curve road.)}
      \label{figure4}
   \end{figure*}

\begin{figure*}[thpb]
      \centering
      \includegraphics[scale = 0.6]{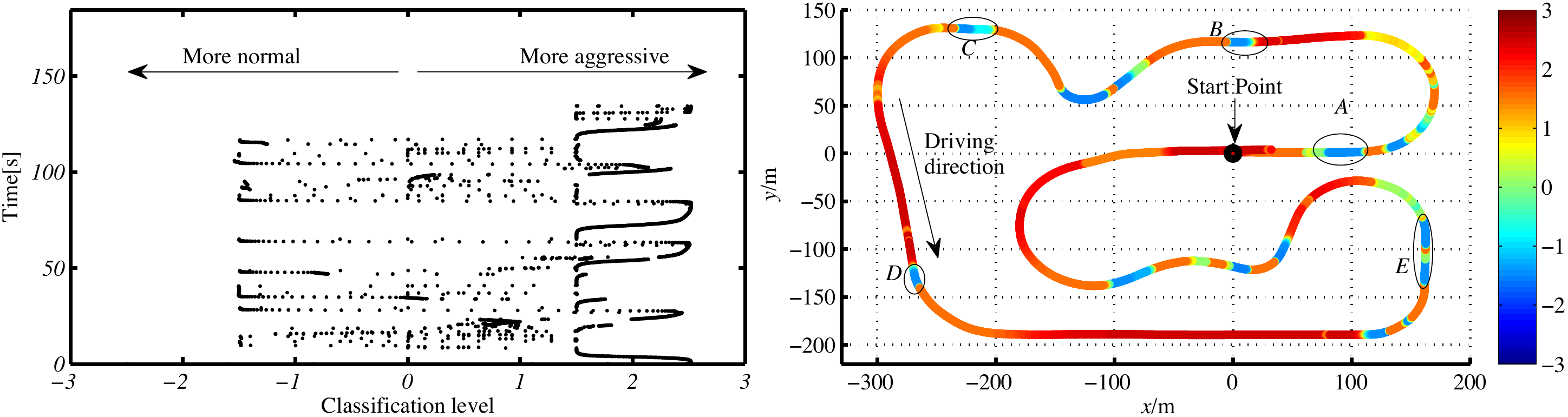}
      \caption{Recognition results for an aggressive driver using the fuzzy logic algorithm. (Left: the classification level; Right: the classification result when driving on the curve road.)}
      \label{figure5}
   \end{figure*}

\begin{figure*}[thpb]
      \centering
      \includegraphics[scale = 0.6]{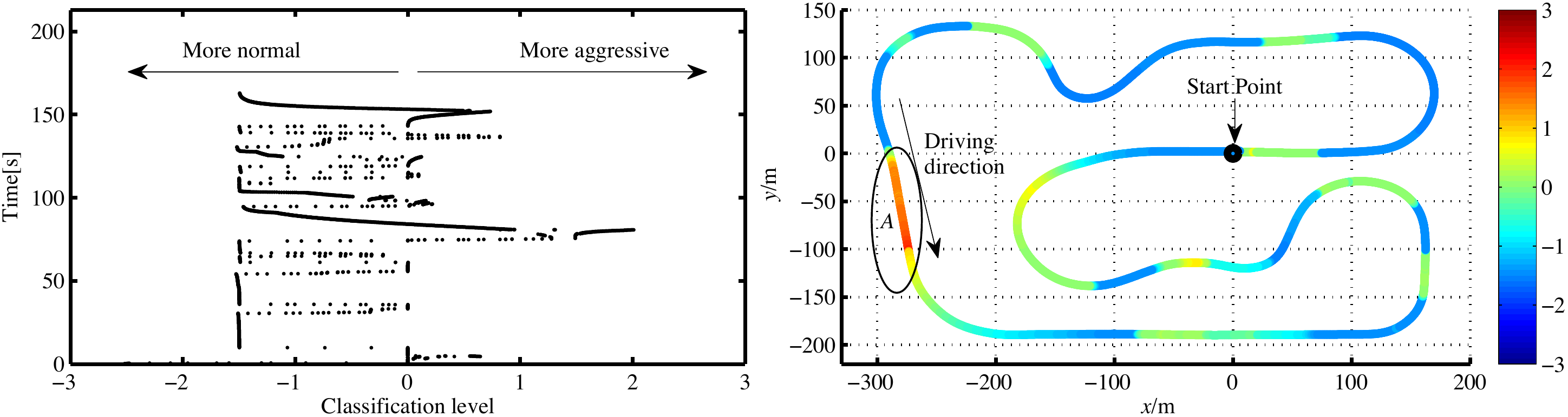}
      \caption{Recognition results for a normal driver using the fuzzy logic algorithm. (Left: the classification level; Right: the classification result when driving on the curve road.)}
      \label{figure6}
   \end{figure*}

For an normal driver, in Fig. \ref{figure4} and \ref{figure6}, driver behaviors are barely labelled as being an aggressive driver. Conversely, most of driving data sets are labelled as being a normal driver.  Furthermore, we found that a normal driver may perform an aggressive driving when driving out of a curve road to a straight line road (shown as part $ A $ in Fig. \ref{figure4} and part $ A $ in Fig. \ref{figure6}), but after entering a curve, the normal driver will perform a normal driving behavior.

\subsubsection{Accuracy Analysis} From Table \ref{Table5}, the average value of $ CRR_{n} $ and $ CRR_{a} $ for eight test drivers (four aggressive and four normal) are $ 0.914 $ and $ 0.862 $, respectively. 

Compared with the FL algorithm, from Table \ref{Table6}, the proposed recognition approach for driving styles is more efficient than the FL algorithm. The proposed algorithm can improve the correctness approximately by $ 3.79 \% $ and $ 22.36 \% $ for aggressive driver and normal driver, respectively. 

\subsubsection{Stability Analysis} From Table \ref{Table6}, for recognition of the normal drivers, the recognition results for using FL algorithm for different normal drivers have a lager difference ($ CRR_{n} $ ranging from $ 0.602 $ to $ 0.870 $), while the recognition results by using the proposed statistical pattern-recognition approach have a smaller difference ($ CRR_{n} $ ranging from $ 0.883 $ to $ 0.980 $), which means that the proposed recognition method is more stable than the FL algorithm. In some degree, the experiment results indicate that the statistical pattern-recognition approach could transform the uncertainty of driver characteristics or driving styles into a relative determinate issue that can be easily overcome.

\begin{table*}[h]
\caption{Recognition results for two types of drivers using the proposed algorithm\label{Table5}}
\begin{center}
\begin{tabular}{c|ccccccc|cc}
\hline
\hline
 Types$ \setminus $Levels   & -3 & -2 & -1 & 0 & 1 & 2 & 3 & $ CRR_a $ & $ CRR_n $\\ 
\hline
\multirow{4}{*}{Aggressive} & 397 &  204 &  292  &  1177 &  332 &   538 &  3485 & 0.861 & - \\
                                  & 467 & 271  & 348  & 830 &  237  &  623  &  3955 & 0.839 & -  \\
                                  & 414 & 220  &  411 &   997 &   462 &  577  & 3506  & 0.841 & - \\
                                  & 158 & 89 &   372 &  1301  &   283 &   869 &  3488 & 0.906 & - \\
\hline
\multirow{4}{*}{Normal} & 4458 & 644 & 626 & 1062 & 257 & 554 & 578 & - & 0.830 \\
                              & 3303  &  1368  &  1064  &  1461   &   419   &  77   &  457 & - & 0.883 \\
                              & 5162  &   1334 &  550   &  1552   &   77    &  129  & 112 & - & 0.964 \\
                              & 3775 &   1823 &  640 &  2193  &  70  &   61  & 37 & - & 0.980  \\
\hline
         Average              &      &        &      &        &      &       &    & 0.862  &  0.914 \\
\hline
\hline
\end{tabular}
\end{center}
\end{table*}

\begin{table}[h]
\begin{center}
\caption{Comparison of recognition results for FL algorithm and the proposed algorithm\label{Table6}}
\begin{tabular}{c|cc}
\hline
\hline
   Type$ \setminus $Algorithm    &  FL  &  Proposed \\
\hline
\multirow{4}{*}{$ CRR_{a} $ of aggressive driver} & 0.812  & 0.861 \\
                            & 0.882  & 0.906 \\
                            & 0.819  & 0.841 \\
                            & 0.809  & 0.839 \\
\hline
Average                & 0.831  & 0.862 \\
\hline
\hline
\multirow{4}{*}{$ CRR_{n} $ of normal driver}     & 0.834  & 0.883 \\
                            & 0.870  & 0.964 \\
                            & \textbf{0.602}   & \textbf{0.980}  \\
                            & \textbf{0.682}   & \textbf{0.914}  \\
                            \hline
 Average               & 0.747                & 0.935 \\                           
\hline
\hline
\end{tabular}
\end{center}
\end{table}

\section{Conclusion}
In this paper, a statistical pattern-recognition method is proposed using kernel density estimation and Euclidean distance to recognize driving styles. This recognition method takes the uncertainty of driving styles into consideration from the viewpoint of statistics. To predict the posterior probability of being fell into which category (aggressive or normal), the full Bayesian theory is involved. To overcome the problem of calculating the covariance among every pair of high dependent elements in feature vector,  the Euclidean distance of projection for each element in feature vector is used for deciding to which category the human driver is belonged, which avoids the complex calculation of covariances among each two elements of feature vectors. And then, a cross-validation method is used to show the benefit of the proposed method, compared with the fuzzy logic algorithm. The recognition results show that the statistical recognition approach for driving styles using kernel density and Euclidean distance could improve the recognition correctness approximately by $ 3.79 \% $ and  $ 22.36 \% $ for aggressive driver and normal driver, respectively, and show a higher stability of recognition, compared with fuzzy logic algorithm.

\section*{Acknowledgment}

The authors would like to thank all the participants who are willing to be the experimental driver for our research and all the members in Vehicle Dynamics \& Control Lab at University of California at Berkeley. This work was supported by China Scholarship Council.

\ifCLASSOPTIONcaptionsoff
  \newpage
\fi



%

%

\begin{IEEEbiography}{Wenshuo Wang}
He received his B.S. in Transportation
Engineering from ShanDong University
of Technology, Shandong, China, in 2012. He is a
Ph.D. candidate for Mechanical Engineering, Beijing
Institute of Technology (BIT). Now he is a visiting
scholar studying in the School of Mechanical Engineering,
University of California at Berkeley (UCB).
Currently, he makes research under the supervisor
of Prof. Junqiang Xi (BIT) and Prof. Karl Hedrick
at Vehicle Dynamics \& Control Lab, University of
California at Berkeley. His research interests include
vehicle dynamics control, adaptive control, driver model, human-vehicle
interaction, recognition and application of human driving characteristics.
His work focuses on modelling and recognising drivers behaviour, making
intelligent control system between human driver and vehicle.
\end{IEEEbiography}

\begin{IEEEbiography}{Junqiang Xi}
He received the B.S. in Automotive
Engineering from Harbin Institute of Technology,
Harbin, China, in 1995 and the PhD in Vehicle
Engineering from Beijing Institute of Technology
(BIT), Beijing, China, in 2001. In 2001, he joined
the State Key Laboratory of Vehicle Transmission,
BIT. During 2012-2013, he made research as an
advanced research scholar in Vehicle Dynamic and
Control Laboratory, Ohio State University(OSU),
USA. He is Professor and Director of Automotive
Research Center in BIT currently. His research interests
include vehicle dynamic and control, powertrain control, mechanics,
intelligent transportation system and intelligent vehicles.
\end{IEEEbiography}

\begin{IEEEbiography}{Xiaohan Li}
He received the M.Sc. degree in
transportation engineering in Beijing Institute of
Technology, Beijing, China, in 2015. Now he is a
PhD candidate in Chair of Human-machine systems
in Technische Universit¨at Berlin, Berlin, Germany.
His group deals with the interaction of drivers
with trendsetting technologies in vehicles. Research
within this scope aims to enhance safety in road
traffic. Special attention is paid to novel concepts for
driver assistance systems, using psychophysiological
data and performance data for implementation and
evaluation.
\end{IEEEbiography}

%
%




\end{document}